\documentclass[final]{cvpr}

\usepackage{times}
\usepackage{epsfig}
\usepackage{graphicx}
\usepackage{amsmath}
\usepackage{amssymb}

\usepackage{amsmath,amssymb}
\usepackage{epsfig}
\usepackage{subfigure}
\usepackage{threeparttable}
\usepackage{booktabs}
\usepackage{caption}

\usepackage[ruled,vlined]{algorithm2e}

\usepackage[pagebackref=true,breaklinks=true,colorlinks,bookmarks=false]{hyperref}

\def\cvprPaperID{20} 
\def\confYear{CVPR 2021}

\begin{document}

\title{Pseudo-IoU: Improving Label Assignment in \\ Anchor-Free Object Detection}

\author{Jiachen Li\textsuperscript{1},
Bowen Cheng\textsuperscript{1},
Rogerio Feris\textsuperscript{2},
Jinjun Xiong\textsuperscript{3},\\
Thomas S. Huang\textsuperscript{1},
Wen-Mei Hwu\textsuperscript{1,4} 
and Humphrey Shi\textsuperscript{1,5,6}\\
\small {\textsuperscript{1}UIUC,
\textsuperscript{2}MIT-IBM Watson AI Lab,
\textsuperscript{3}IBM T.J. Watson Research Center,}\\
\small {\textsuperscript{4}NVIDIA,
\textsuperscript{5}University of Oregon,
\textsuperscript{6}Picsart AI Research (PAIR)}}

\maketitle
\begin{abstract}
 Current anchor-free object detectors are quite simple and effective yet lack accurate label assignment methods, which limits their potential in competing with classic anchor-based models that are supported by well-designed assignment methods based on the Intersection-over-Union~(IoU) metric. In this paper, we present \textbf{Pseudo-Intersection-over-Union~(Pseudo-IoU)}: a simple metric that brings more standardized and accurate assignment rule into anchor-free object detection frameworks without any additional computational cost or extra parameters for training and testing, making it possible to further improve anchor-free object detection by utilizing training samples of good quality under effective assignment rules that have been previously applied in anchor-based methods. 
By incorporating Pseudo-IoU metric into an end-to-end single-stage anchor-free object detection framework, we observe consistent improvements in their performance on general object detection benchmarks such as PASCAL VOC and MSCOCO. Our method (single-model and single-scale) also achieves comparable performance to other recent state-of-the-art anchor-free methods without bells and whistles.
Our code is based on mmdetection toolbox and will be made publicly available at \href{https://github.com/SHI-Labs/Pseudo-IoU-for-Anchor-Free-Object-Detection}{https://github.com/SHI-Labs/Pseudo-IoU-for-Anchor-Free-Object-Detection}.
\end{abstract}

\section{Introduction}

\begin{figure}[htb]
\centering
\includegraphics[width=0.45\textwidth]{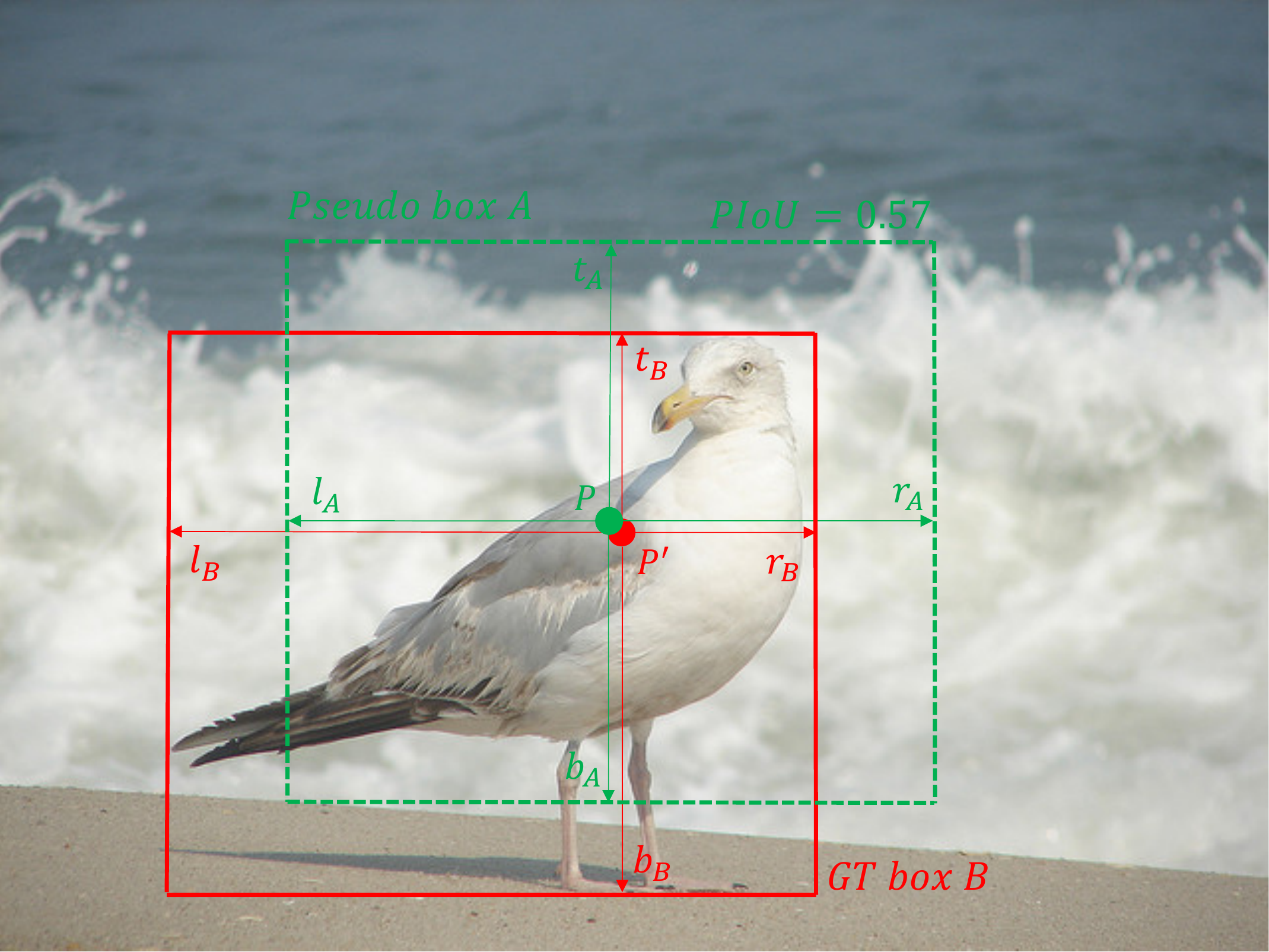}
\caption{As shown in the image, the red box is the ground truth box and the green box is a pseudo box with the same size to the ground truth box. The green point $P$ and red point $P^{'}$ are the same point from feature map projection. $(l,r,t,b)$ represent distances from a point inside box to the left, right, top and bottom side of the box, respectively. Point $P$ is at center of the pseudo box $A$ and Point $P^{'}$ is in the ground truth box $B$. The Pseudo-IoU between the two boxes is 0.57.}
\label{Figure 1}
\end{figure}

Label assignment is very important in  training accurate object detection models. In recent years, anchor-based object detection methods
have been popular in the community, which starts with Faster RCNN~\cite{ren2015faster}. Typically, during the training process, after feature extraction with deep convolutional neural networks~(DCNN), each point on the feature maps is assigned with multiple anchors and each anchor is assigned a positive or negative label based on its overlap with ground truth provided from the dataset.
Then, proportional positive and negative samples are selected for training the network and the whole process is defined as label assignment in anchor-based object detection. Since the assignment process selects training samples for regression and classification, improving assignment method could significantly enhance detection results by assigning accurate training samples. Recent work on Cascade RCNN~\cite{cai2018cascade} employs cascaded classifiers and regressors with improved  Intersection-over-Union~(IoU) thresholds to assign better positive training samples. OHEM~\cite{shrivastava2016training} adopts online hard negative sampling to bring more hard negative samples with larger losses into training. Libra RCNN~\cite{pang2019libra} uses an IoU-based sampling method to achieve a more balanced training. All these prior works prove that better assignment for training samples is indispensable and effective in anchor-based object detection.

Recently, another popular branch of object detection methods are anchor-free models that do not assume pre-defined anchors during the whole training process, which reduces many hyper-parameters with anchors that require heuristic tuning for good performance. Anchor-free models predict bounding boxes directly from points to the left, right, top and bottom side of ground truth box like FCOS~\cite{tian2019fcos} and FSAF~\cite{Zhu_2019_CVPR}. However, due to lack of accurate assignments, they both use other methods to compensate for the performance gap. For FCOS, it views all points inside shrinked ground truth as positive samples and adds a centerness branch to reweigh detection outputs that decreases some false positives. For FSAF, it employs online feature selection and a combination of anchor-free and anchor-based methods.

In this paper, we propose a Pseudo-Intersection-over-Union~(Pseudo-IoU) metric that brings an accurate label assignment rule into current anchor-free object detectors, which is illustrated in Figure~\ref{Figure 1}. For each feature map point inside ground truth boxes, after mapping to the original input image, we assume a corresponding pseudo box that is centered at the point and has the same size to ground truth box.
Then we can easily compute IoU between the centered pseudo box and ground truth box. Since the IoU is based on a pseudo box assigned to each point, we name the metric Pseudo-Intersection-over-Union~(Pseudo-IoU). After Pseudo-IoU computation, each point can be assigned a Pseudo-IoU value $v$ like each anchor with an IoU for assignment in anchor-based methods. Now each point on the feature map pyramid has a Pseudo-IoU value $v \in [0,1]$, they can be labeled as positive or negative training samples based on a Pseudo-IoU threshold $T \in [0,1]$ with,

\vspace{-1mm}
$$
labels = \left\{
\begin{aligned}
+1 ~~~~& if & v \geq T\\
-1 ~~~~& if & v < T \\
\end{aligned}
\right.
$$
\vspace{-1mm}

From Figure~\ref{Figure 1}, it shows some points near the sides of bounding boxes are assigned as negative samples. It leads to more false positives and inaccurate bounding boxes if taking these points with low Pseudo-IoUs as positive samples. To demonstrate the effectiveness of our Pseudo-IoU metric, we build an anchor-free baseline with ResNet-101~\cite{he2016deep} as backbone and FPN~\cite{lin2017feature} as neck~\cite{chen2019mmdetection} to make dense anchor-free predictions on enhanced feature map pyramids. Furthermore, we add Pseudo-IoU metric for assignment and conduct extensive experiments on Pascal VOC and MSCOCO dataset. Specifically, the anchor-free methods with Pseudo-IoU based assignment outperforms baseline by 2.1\% mean Average Precision~(mAP) on Pascal VOC 2007 test set, 3.0\% mAP on MSCOCO minival set and 3.1\% mAP on MSCOCO test-dev set. It also reaches results comparable to recent state-of-the-art methods without bells and whistles.

To summarize, our contributions are as follows:
\begin{itemize}
\item We investigate and analyze the problems and bottlenecks of current anchor-free object detectors, which is lack of accurate label assignment process that is well-defined and designed in anchor-based methods.
\item We propose Pseudo-Intersection-over-Union~(Pseudo-IoU) metric, which takes accurate label assignment rule into anchor-free framework without any additional cost during training or testing, making it possible to improve anchor-free detection by improving assignment process.
\item Our single-scale model improves the performance of baseline by a large margin and achieves performance very comparable to other recent state-of-the-art detection methods.

\end{itemize} 

\section{Related Work}

\noindent \textbf{Anchor-based Object Detectors.}~In the past few years, anchor-based object detectors have been the mainstream in the object detection area. It starts with Faster R-CNN~\cite{ren2015faster}, a supreme version of previous R-CNN~\cite{girshick2014rich} and Fast R-CNN~\cite{girshick2015fast}, which first proposes the idea of using anchors as pre-defined bounding boxes for subsequent regression to ground truth. Following Faster RCNN, most two-stage object detectors inherit the anchor-based manner, like  FPN\cite{lin2017feature}, Mask RCNN~\cite{he2017mask}, Cascade RCNN~\cite{cai2018cascade}, DCR~\cite{cheng2018decoupled},
GFR~\cite{shen2017improving} and other methods~\cite{shi2018geometry}~\cite{zhang2019skynet}. Some methods adopt efficient training like SNIP~\cite{singh2018analysis} and SNIPER~\cite{singh2018sniper}, that take clips of images for training. Some approaches improve quality of anchors like GA-RPN~\cite{wang2019region}, which learn the setting of anchors for efficient training. Moreover, some popular one-stage object detectors also employ anchors for better regression, which are proposal free and directly make predictions from anchors to final bounding boxes, like SSD~\cite{liu2016ssd}, YOLOv2~\cite{redmon2016you}, RetinaNet~\cite{lin2017focal}, RefineDet~\cite{zhang2018single} and GHM~\cite{li2019gradient}, introduces gradient harmonized loss function for a balanced training process.\\

\noindent \textbf{Anchor-Free Object Detectors.}~Different from anchor-based object detectors, another way to predict bounding boxes is to regress directly from points, which starts from Densebox~\cite{huang2015densebox} and YOLO~\cite{redmon2016you}. They make prediction by regressing the bounding boxes directly from central points in ground truth area to final bounding boxes. Moreover, with the power of feature pyramid networks, recent paper like FCOS~\cite{tian2019fcos} and FSAF~\cite{Zhu_2019_CVPR} build a whole anchor-free detection pipeline, which reach results very comparable to one-stage object detectors like RetinaNet. Another branch is keypoint-based anchor-free object detection, which makes bounding boxes prediction after keypoint prediction, including Cornernet~\cite{law2018cornernet}, Centernet~\cite{zhou2019objects} and Extremenet~\cite{zhou2019bottom}. Since the key point prediction is sparse and accurate, post processing like NMS can be removed in these methods.\\
\\
\noindent \textbf{Assignment Rules.}~Recently, researchers notice that label assignment is essential for a balanced and stable training process. For anchor-based object detectors, anchors are assigned with positive or negative labels according their IoU with ground truth. To improve quality of positive samples, Cascade RCNN~\cite{cai2018cascade} employs increased IoU threshold to select positive samples to train cascaded classifiers. OHEM~\cite{shrivastava2016training} uses a hard false positive mining branch which select more samples with higher losses as negative training samples. Libra RCNN~\cite{pang2019libra} further proposes an IoU based sampling methods and PISA~\cite{cao2019prime} focuses on prime samples for training. ATSS~\cite{zhang2019bridging} focuses on bridging gap between anchor-free and anchor-based methods with adaptively selecting positive and negative samples according to statistical characteristics of objects. SAPD~\cite{zhu2019soft} uses soft anchor point for label assignment and AutoAssign~\cite{zhu2020autoassign} employs automatic label assignment strategy in anchor-free detectors. 

\section{Our Approach}

In this section, we firstly review assignment process at anchor-based detectors, then propose how we employ Pseudo-IoU metric in anchor-free detectors to make accurate assignment available during training process. Next, we show the whole pipeline of our anchor-free detector with Pseudo-IoU based assignment and introduce each component in the pipeline.

\subsection{Pseudo-Intersection-over-Union}
Intersection-over-Union~(IoU) has been well-defined and applied in anchor-based methods. It is used for comparing similarity between two arbitrary shapes $A$ and $B$, and the IoU between them is, 
$$
IoU (A,B) = \frac{|A \cap B|}{|A \cup B|} 
=\frac{|A \cap B|}{|A|+|B|-|A \cap B|}
$$

\begin{algorithm}[tb]
\SetAlgoLined
\textbf{1.}~ 
Get ~$l_B$, $r_B$, $t_B$, $b_B$, $l_A$, $r_A$, $t_A$, $b_A$, $S_A$ and $S_B$: \\
\begin{center}
$l_A = r_A = (l_B+r_B)/2$ \\
$t_A=b_A=(t_B+b_B)/2$ \\
$S_B = (l_B+r_B)*(t_B+b_B)$ \\
$S_A = (l_A+r_A)*(t_A+t_A)$ \\
\end{center}
 \textbf{2.}~
Intersection box parameters:
\begin{center}
 $l_* = min(l_A,l_B)$ \quad $r_* = min(r_A,r_B)$\\
$t_* = min(t_A,t_B)$ \quad $b_* = min(b_A,b_B)$
\end{center}
 \textbf{3.}~
Intersection box area:
\begin{center}
 $|A\cap B| = S_* = (l_*+r_*)*(t_*+b_*)$
\end{center}
\textbf{4.}~
Union box area:
\begin{center}
 $|A\cup B| = S_A+S_B-S_*$
\end{center}
\textbf{5.}~
Compute Pseudo-IoU:
\begin{center}
 $Pseudo-IoU = \frac{|A\cap B|}{|A\cup B|} = \frac{S_*}{S_A+S_B-S_*} $
\end{center}
\caption{Pseudo-Intersection-over-Union}
\end{algorithm}

In anchor-based object detectors like Faster RCNN~\cite{ren2015faster}, anchors are pre-defined bounding boxes surrounding with the center of sliding windows, namely points on feature maps. 
For an anchor $A$ with an area $S_A$ and a ground truth $B$ with an area $S_B$, their IoU will be
$$
IoU (A,B) = \frac{|S_A \cap S_B|}{|S_A \cup S_B|} 
=\frac{|S_A \cap S_B|}{|S_A|+|S_B|-|S_A \cap S_B|}
$$
\begin{figure*}[htb]
\centering
\includegraphics[width=1\textwidth]{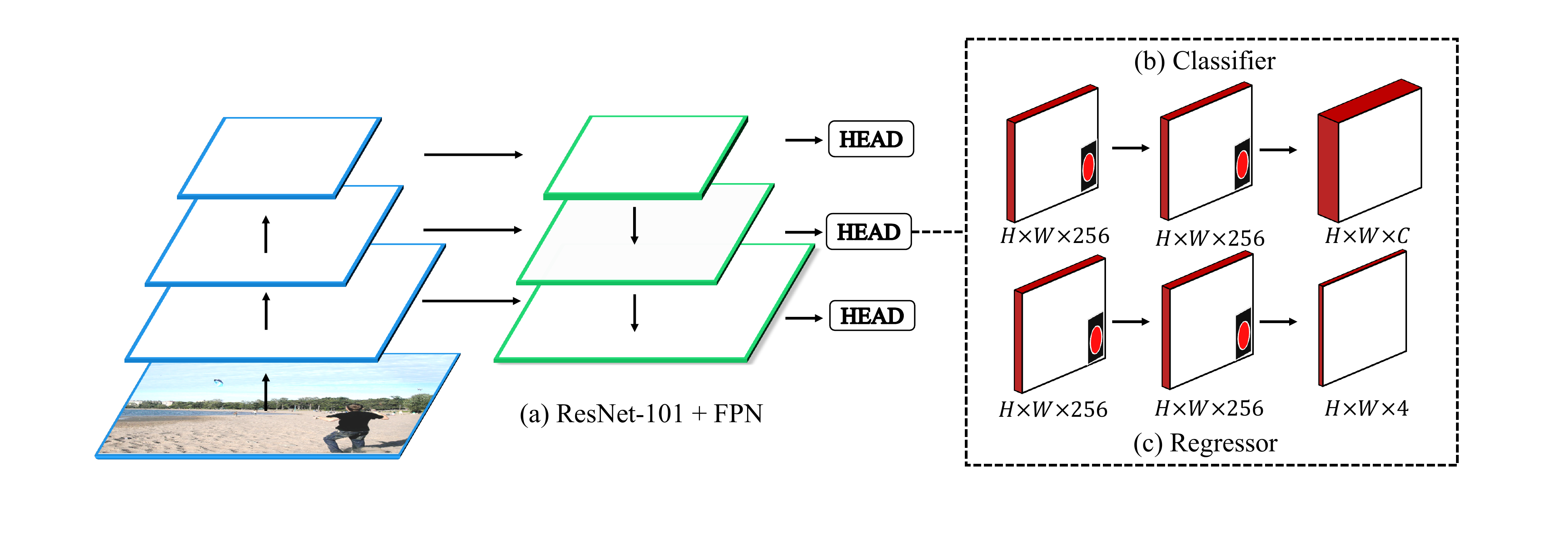}
\caption{Anchor-free object detector pipeline with Pseudo-IoU based assignment.} 
\label{Figure 3}
\end{figure*}

Typically, an anchor has an IoU overlap more than $0.5$ would be assign as a positive sample and has an IoU overlap less than $0.4$ would be assign as a negative sample. This IoU-based process is adopted by most of two-stage detectors and some one-stage detectors like SSD~\cite{liu2016ssd} and RetinaNet~\cite{lin2017focal}. However, for recent anchor-free object detectors, there is no such counterpart during training process. 
The main reason is that without anchors, the regressor could not regress offsets from anchors to ground truth boxes. On the contrary, it regresses from a point inside ground truth to the four sides to make direct predictions. For example, given a ground truth box $B$ and a point $P$ from feature map that is inside the box,  $l_B$, $r_B$, $t_B$ and $b_B$ represent distance from point $P$ to the left, right, top and bottom side of box $B$. we can view the point $P$ as a center point and assign a pseudo box $A$ around it,  $l_A$, $r_A$, $t_A$ and $b_A$ represent distance from point $P$ to the left, bottom, right and top side of pseudo box $A$. Then the process to compute IoU between ground truth box $B$ and pseudo box $A$ is described in $Algorithm~ 1$.

We name the IoU between the assigned pseudo box and ground truth box Pseudo-Intersection-over-Union, namely Pseudo-IoU, and the valve of Pseudo-IoU is $v \in [0,1]$, making it possible for selecting training examples with a specific Pseudo-IoU threshold in anchor-free pipelines.

\subsection{Anchor-Free Architecture}
Most state-of-the-art anchor-free detectors follow the architecture of RetinaNet, which uses a ResNet-101~\cite{he2016deep} as backbone and FPN~\cite{lin2017feature} as neck~\cite{chen2019mmdetection}, followed by detection heads with two branches: one for image classification with focal loss function, and the other one for bounding boxes regression with smooth L1 loss function. Our anchor-free detection architecture with Pseudo-IoU based assigner is illustrated in Figure~\ref{Figure 3}.\\

\noindent \textbf{Detection Backbone and Neck}~
Following RetinaNet, we adopt ResNet-101 as our backbone to extract features and use FPN to augment feature maps through top-down pathway and lateral connections, which bulids a rich multi-scale feature pyramids with a single resolution input image and has been proved a strong detection component, as we illustrated in Figure~\ref{Figure 3} Part(a). Similar to RetinaNet, in the downsampling period, we use 5 stages in ResNet that generates feature maps $C_1$, $C_2$, $C_3$, $C_4$, $C_5$ and the downsampling rate is $2^l$ for $C_l$ feature map. In the upsampling stage, we remove $P_2$ for less computational burden and use feature maps $P_3$, $P_4$, $P_5$, $P_6$ and $P_7$. $P_3$, $P_4$ and $P_5$ are generated from $C_3$, $C_4$ and $C_5$ through upsampling and lateral connections. $P_6$ and $P_7$ are computed by stride-2 convolution over $P_5$ for detection on large targets. All feature maps from $P_3$ to $P_7$ have an output channel $C=256$, which are sent to detection heads for subsequent training process.\\

\noindent \textbf{Pseudo-IoU based Assignment}~
After feature extraction by ResNet-101 and FPN, we can assign points in feature maps from $P_3$ to $P_7$ with positive or negative labels. For a point at feature map $P_l$, its location can be projected back to original input image with an upsampling rate $2^l$. Suppose that its location is $(x,y)$ on the input image and if it falls into a ground truth box with label $(x_c,y_c,w,h)$, $(x_c,y_c)$ is the center of the box and $w$, $h$ represent its width and height respectively. Then, we can compute $l$, $r$, $t$, $b$, which are distance from $(x,y)$ to the left, right, top and bottom side of box $(x_c,y_c,w,h)$ by 
\begin{center}
$ l = x - x_c + \frac{w}{2} ,\quad r = x_c + \frac{w}{2} - x $
\end{center}
\begin{center}
$ t = y - y_c + \frac{h}{2} ,\quad b = y_c + \frac{h}{2} - y $
\end{center}
according to the Pseudo-IoU computation algorithm in section $3.1$, we can easily have
\begin{center}
$l_B = l, r_B=r, t_B=t, b_B=b$ 
\end{center}
then compute Pseudo-IoU following the flow diagram presented in $Algorithm~1$. In anchor-based methods, it usually takes anchors with IoU more than a specific threshold as positive samples, we follow the design and set our Pseudo-IoU threshold $T$, too. Therefore, all points with Pseudo-IoU larger than $T$ are assigned as positive samples and others are assigned as negative samples. Since focal loss performs good and stable on balanced training and FCOS\cite{tian2019fcos} further proves that one main advantage of anchor-free detectors is all-in samples for training, we use all labeled samples for subsequent training. In Figure~\ref{Figure 3}, for example, on all $H\times W \times 256$ feature maps, there is a black rectangle ground truth area and the red area inside contains all positive samples contribute to the training procedure.\\

\noindent \textbf{Detection Heads}~ After all feature maps are extracted from FPN and all points on the feature maps are labeled with positiveness or negativeness, the training process moves forward to detection heads part. The detection head is a FCN~\cite{long2015fully} that attached to each output feature maps from FPN and it contains two subsets: a classifier and a regressor. For the classifier subnet illustrated in Figure~\ref{Figure 3} part(b), it follows four stacked $3 \times 3$ convolution layers with 256 filters and finally attached a $3 \times 3$ convolution layer with $C$ filters. $C$ is the number of classes for final classification on each point. For the regressor subnet, in Figure~\ref{Figure 3} part(c), it also follows four stacked $3 \times 3$ convolution layers with 256 filters and finally attached a $3 \times 3$ convolution layer with 4 filters per spatial location. At each location, these 4 outputs represent a predicting bounding box vector $(l,r,t,b)$. Moreover, the two subsets do not share weights on the four consecutive stacked $3 \times 3$ convolution layers.\\ 

\begin{table*}[tb]
\begin{minipage}{.31\textwidth}
\centering
\begin{tabular}{c|c|c}
\toprule[2pt]
Model  &Thrs   & mAP   \\ 
\midrule[1.5pt]
Baseline &0.0 &76.9 \\ \hline
Pseudo-IoU &0.1  &76.8 \\ 
- &0.2  & 77.5\\
- &0.3  &78.2 \\
- &\textbf{0.4}  &  \textbf{79.0} \\
- &0.5     & 78.9 \\
- &0.6  & 45.6 \\
- &0.7        & 22.1      \\  \hline
\bottomrule[2pt]
\end{tabular}
\caption*{(a)~Diff Pseudo-IoU Thrs}
\end{minipage}
\begin{minipage}{.32\textwidth}
\centering
\begin{tabular}{c|c|c}
\toprule[2pt]
Model& Epochs     & mAP   \\ 
\midrule[1.5pt]
Baseline & 8  &76.9 \\ 
Pseudo-IoU & 8  &79.0 \\  \hline
Baseline &12  &78.4 \\
Pseudo-IoU & \textbf{12}  &\textbf{80.4} \\  \hline
Baseline &16  &78.3 \\ 
Pseudo-IoU & 16  &80.2 \\  \hline
Baseline &24  &77.8 \\ 
Pseudo-IoU & 24  &80.1 \\  \hline
\bottomrule[2pt]
\end{tabular} 
\caption*{(b)~Diff Training Epos}
\end{minipage}
\begin{minipage}{.32\textwidth}
\centering
\begin{tabular}{c|c|c}
\toprule[2pt]
Model& Batchsize & mAP   \\ 
\midrule[1.5pt]
Baseline & 4  & 73.6\\ 
Pseudo-IoU & 4  & 75.2\\  \hline
Baseline & 6  &76.6 \\ 
Pseudo-IoU & 6  & 77.9\\  \hline
Baseline & 8 & 77.0\\
Pseudo-IoU & 8  & 78.8\\  \hline
Baseline &16  &76.9 \\ 
Pseudo-IoU & 16  &\textbf{79.0} \\  \hline
\bottomrule[2pt]
\end{tabular} 
\caption*{(c)~Diff Batchsizes}
\end{minipage}
\caption{Ablation studies results evaluated on Pascal VOC 2007 test set. \textbf{Bold} indicates best single-model results and corresponding parameters.} 
\label{Table 1}
\end{table*}

\begin{table}[htb]
\centering
\begin{tabular}{c|c|c|c|c|c}
\toprule[2pt]
Centerness    & mAP   &S-box  & mAP  & P-IoU & mAP  \\ 
\midrule[1.5pt]
0.2  &72.9 & 0.3 &24.5 & 0.3 &78.2\\ \hline
0.25  & 77.4& 0.4 &63.8 & \textbf{0.4} &\textbf{79.0}\\ \hline
0.3  & 78.5& 0.5 &77.1 & 0.5 &78.9\\ \hline
\textbf{0.35}     &\textbf{78.7}& \textbf{0.6} &\textbf{78.3}& 0.6 &45.6 \\ \hline
0.4       & 77.6 & 0.7 &77.3 & 0.7 &22.1\\ \hline
0.45      & 77.6  & 0.8 &76.8 & 0.8 &13.5\\ \hline
\bottomrule[2pt]
\end{tabular}

\caption{Ablation studies with different assignment rules and all results evaluated on Pascal VOC 2007 test set. \textbf{Bold} indicates best single-model results and corresponding parameters.} 
\label{Table 2}
\end{table}

\noindent \textbf{Loss Function}~ 
The loss function in our anchor-free architecture is
$$
L(A_{(x,y)}, B_{(x,y)}) = \frac{1}{N_{pos}} \sum_{(x,y)} L_{fl}(C_{{A_{x,y}}},C_{{B_{x,y}}})
$$
$$
+  \frac{\lambda}{N_{pos}} \sum_{(x,y)}  L_{iou}(R_{{A_{x,y}}},R_{{B_{x,y}}})
$$
where $L_{fl}$ is focal loss for classification and $L_{iou}$ is IoU loss~\cite{yu2016unitbox} for regression. $A_{(x,y)}$ is the set of all selected training samples and $B_{(x,y)}$ is the set of all ground truth boxes. $C_{{A_{x,y}}}$ and $R_{{A_{x,y}}}$ represent the output of classifier and regressor. $C_{{B_{x,y}}}$ and $R_{{B_{x,y}}}$ represent structured labels of ground truth. $N_{pos}$ is the number of selected positive samples and $\lambda $ denotes the balance weight for IoU loss.\\

\noindent \textbf{Inference Process}~ 
The inference process is straightforward and clear for anchor-free detectors. The input image goes through backbone ResNet-101 and FPN to generate feature maps from $P_3$ to $P_7$, then the regressor and classifier make predictions on all points and assemble them together for post processing. Generally, the post processing includes a confidence thresholding and a non maximum suppression~(NMS) for final predictions.

\section{Experiments}
We present all of our experiments mainly on Pascal VOC~\cite{everingham2010pascal} and MSCOCO~\cite{lin2014microsoft} benchmark. For Pascal VOC benchmark, we follow common $07+12$ practice that uses Pascal VOC 2007 trainval split and 2012 trainval split for training and 2007 test split for testing. For MSCOCO benchmark, we use trainval35k for training, common minival 5k images for validating and test on test-dev set. We first show the training and testing setting of the two benchmarks. Then, we present the effectiveness of Pseudo-IoU in ablation study and compare our model with other state-of-the-art detectors on MSCOCO dataset. All the codes are based on mmdetection~\cite{chen2019mmdetection} toolbox.

\subsection{Training and Testing}
\noindent \textbf{Pascal VOC}~ 
We use an ImageNet~\cite{deng2009imagenet} pretrained ResNet-101\cite{he2016deep} as backbone with fixed Batch Normalization~\cite{ioffe2015batch} layers and build the whole network illustrated in Figure~\ref{Figure 3} with number of classes $C = 21$ and Group Normalization~\cite{wu2018group} for stable training. Our network is trained with stochastic gradient descent (SGD) for 12 epochs with a warmup\cite{goyal2017accurate} training for 500 iterations. The initial learning rate is 0.01 and reduced by a factor of 10 at epoch 8 and 10, respectively. Weight decay and momentum are set to be 0.0001 and 0.9. The input images size are resized to have their shorter side being 600 and their longer side less or equal to 1000. We only use image flipping as the only data augmentation method since most experiments on Pascal VOC are ablation studies. The model is trained on 4 $\times$ 1080 Ti GPU with a total batch size at 16. The loss function is an addition of focal loss and IoU loss, the weight balance parameter of IoU loss is $\lambda = 1$. In the testing process, we only use same single resolution to the training process for inference. For post-processing, the scoring threshold and NMS's threshold are set to be 0.05 and 0.5, respectively.  \\

\begin{table*}[t]
\centering
\begin{tabular}{c|ccc|ccc|ccc}
\toprule[2pt]
Method  &Pseudo-IoU &GIoU &Centerness    &$AP$ &$AP^{50}$ &$AP^{75}$ &$AP^S$  &$AP^M$  &$AP^L$  \\ 
\midrule[1.5pt]
RetinaNet\cite{lin2017focal} &- &- &- &35.7 &55.0 &38.5 &18.9 &38.9 &46.3 \\ \hline
AF-Baseline & - &- &-&33.9 &54.5 &35.3 &19.1 &38.7 &44.1 \\ \hline
Pseudo-IoU  & \checkmark & - & - &36.8 &56.4 &39.2 &20.4 &41.0 &48.7\\ \hline
Pseudo-IoU & \checkmark & \checkmark & - &37.1 &55.9 &39.7 &20.4 &41.1 &49.1\\ \hline
Pseudo-IoU  & \checkmark & \checkmark & \checkmark &37.4 &56.5 &40.1 &20.5 &41.2 &49.4\\ \hline

\bottomrule[2pt]
\end{tabular} 
\caption{Anchor-free model with Pseudo-IoU based assignment on MSCOCO minival 5k set with ResNet-50-FPN backbone. AF-baseline stands for anchor-free baseline. Pseudo-IoU could bring $2.9\%$ mAP improvement on baseline model. Using GIoU\cite{rezatofighi2019generalized} for regression loss and adding a centerness branch like FCOS\cite{tian2019fcos} could both bring further improvements.}
\label{Table 3}
\end{table*}

\noindent \textbf{MSCOCO}~ 
We build the whole network with a illustrated in Figure~\ref{Figure 3} with the number of classes $C = 81$ and Group Normalization~\cite{wu2018group} for stable training. Our network employs ResNet-101 as backbone with feature pyramid networks and is trained with stochastic gradient descent (SGD) for 24 epochs with a warmup training for 1500 iterations. The initial learning rate is 0.01 and reduced by a factor of 10 at epoch 16 and 22, respectively. Weight decay and momentum are set to be 0.0001 and 0.9. The input images are resized to have their shorter side being 800 and their longer side less or equal to 1333 for single-scale training. The model is trained on 4 $\times$ V100 GPU with a total batch size at 16. The loss function setting remains unchanged. During the testing process, we only use same single resolution to the training settings for inference with the same post-processing used at Pascal VOC's models.

\begin{table*}[htb]
\centering
\begin{tabular}{c|c|ccc|ccc}
\toprule[2pt]
Method     &Backbone  &$AP$ &$AP^{50}$ &$AP^{75}$ &$AP^S$  &$AP^M$  &$AP^L$  \\ 
\midrule[1.5pt]
\textbf{Multi-Stage methods} \\ \hline
Faster RCNN~\cite{ren2015faster} & R-101-FPN &36.2 &59.1 &39.0 &18.2 &39.0 &48.2 \\ 
R-FCN~\cite{dai2016r} & Aligned-Inception-R &35.5 &55.6 &- &17.8 &38.4 &49.3 \\
Deformable R-FCN~\cite{dai2017deformable} & Aligned-Inception-R &37.5 &58.0 &-  &19.4 &40.1 &52.5 \\
Mask RCNN~\cite{he2017mask} &R-101-FPN &38.2 &60.3 &41.7 &20.1 &41.1 &50.2 \\
Libra RCNN~\cite{pang2019libra} &R-101-FPN &41.1 &62.1 &44.7 &23.4 &43.7 &52.5 \\
Cascade RCNN~\cite{cai2018cascade}&R-101-FPN &42.8 &62.1 &46.3 &23.7 &45.5 &55.2 \\\hline
\textbf{One-Stage methods} \\ \hline
YOLOv3~\cite{redmon2018yolov3}  &Darknet-53 &33.0 &57.9 &34.4 &18.3 &25.4 &41.9 \\
SSD513~\cite{liu2016ssd} &R-101 &31.2 &50.4 &33.3 &10.2 &34.5 &49.8 \\
DSSD513~\cite{fu2017dssd}  &R-101 &33.2 &53.3 &35.2 &13.0 &35.4 &51.1 \\
RefineDet~\cite{zhang2018single}   &R-101 &36.4 &57.5 &39.5 &16.6 &39.9 &51.4 \\
RetinaNet~\cite{lin2017focal}  &R-101-FPN &39.1 &59.1 &42.3 &21.8 &42.7 &50.2 \\
GHM ~\cite{li2019gradient}   &R-101-FPN &39.9 &60.8 &42.5 &20.3 &43.6 &54.1 \\
  \hline
\textbf{Anchors-Free methods} \\ \hline
FSAF~\cite{Zhu_2019_CVPR} &R-101-FPN &40.9 &61.5 &44.0 &24.0 &44.2 &51.3 \\
FCOS~\cite{tian2019fcos} &R-101-FPN &41.5 &60.7 &45.0 &24.4 &44.8 &51.6 \\ 
CornetNet~\cite{law2018cornernet} &Hourglass-104 &40.5 &56.5 &43.1 &19.4 &42.7 &53.9 \\ 
CenterNet~\cite{zhou2019objects} &Hourglass-104 &42.1 &61.1 &45.9 &24.1 &45.5 &52.8 \\ \hline
\textbf{Our methods} \\ \hline
AF-Baseline &R-101-FPN &38.3 &59.2 &40.1 &22.6 &42.1 &46.9 \\
Pseudo-IoU &R-101-FPN &41.5 &61.0 &44.5 &24.1 &44.6 &51.9\\
Pseudo-IoU &R-101-FPN-DCN &43.4 &63.2 &46.8 &25.8 &47.8 &56.7\\ 
Pseudo-IoU &R-32x8d-FPN-DCN &44.0 &63.8 &47.3 &28.0 &47.5 &58.1\\ 
Pseudo-IoU &R-64x4d-FPN-DCN &44.5 &64.4 &48.1 &26.5 &48.8 &58.4\\ \hline

\bottomrule[2pt]
\end{tabular} 
\caption{Detection results of our best single model with Pseudo-IoU based label assignment vs. state-of-the-art one-stage, multi-stage and anchor-free detectors on the MSCOCO test-dev set. R stands for ResNet here.}
\label{Table 4}
\end{table*}

\subsection{Ablation Study}
\noindent \textbf{Pseudo-IoU is Essential}~  
In Table~\ref{Table 1}, we carefully present all ablation studies experiments with the effectiveness of Pseudo-IoU on Pascal VOC 2007 test set. The baseline model is in Figure 3 without Pseudo-IoU based assignment, which takes all points inside ground truth as positive samples. Table~\ref{Table 1}.(a) shows that the baseline model with a 0.4 Pseudo-IoU threshold could get a 2.1\% improvement on mAP, which proves the importance of accurate assignment during training process. It also shows that a higher Pseudo-IoU threshold would lead to a performance drop due to insufficient selected positive training samples. In Table~\ref{Table 1}.(b) and (c), it indicates that anchor-free model with Pseudo-IoU based assignment could bring a consistent around 2\% mAP improvement under different training epochs and batchsizes, which shows that the performance gap due to imbalanced assignment of training samples could not be compensated with extended training and larger batchsize.\\

\noindent \textbf{Assignment with Other Metrics}~ 
In current state-of-the-art anchor-free object detectors, FCOS~\cite{tian2019fcos} and FSAF~\cite{Zhu_2019_CVPR} also achieve detection results comparable to anchor-based methods. In FCOS, it adds a centerness branch to reweigh output results. In FSAF, it adopts a central part in ground truth box as positive samples. These two recent work motivate us to bring a Pseudo-IoU based assignment method into anchor-free pipeline. Moreover, there are some other assignment options may be as effective as Pseudo-IoU. We propose other two options adapted from FSAF and FCOS that may be alternatives to our Pseudo-IoU metric. The first one is called scaled-box, which uses all points inside scaled ground truth boxes as positive samples. Given a point $(x,y)$ that falls into a ground box $(x_c,y_c,w,h)$ and a scaled parameter $s$, the point is positive if it fits that
$$
|x-x_c|<\frac{sw}{2} \quad |y-y_c|<\frac{sh}{2}
$$
The other one is called centerness, under the same setting to scaled-box, we can compute $l,r,t,b$ according to algorithm presented in section 3.2, then centerness threshold is 
$$
c = \sqrt{\frac{\min(l,r)}{\max(l,r)} \times \frac{\min(t,b)}{\max(t,b)}}
$$
which follows the expression of centerness in FCOS. However, here we use centerness for selecting positive samples rather than reweigh output detection. We compare the detection results on Pascal VOC 2007 test set under these two new metrics and Pseudo-IoU in Table~\ref{Table 2}. It shows that anchor-free models can achieve different results under different assignment methods. Overall, using Pseudo-IoU could achieve a little better results (79.0\% mAP) than centerness (78.7\% mAP) and scaled-box (78.3\% mAP) metric. More importantly, use Pseudo-IoU based metric may employ previous research on IoU like Cascade RCNN~\cite{cai2018cascade} or Libra RCNN~\cite{pang2019libra} into anchor-free manners, which bring more possibilities to anchor-free object detection.

\begin{figure*}[htb]
\centering
\vspace{-.10in}
\subfigure{
\centering
\includegraphics[width=0.191\textwidth,height=0.11\textheight]{10_baseline.jpg}}
\subfigure{
\centering
\includegraphics[width=0.191\textwidth,height=0.11\textheight]{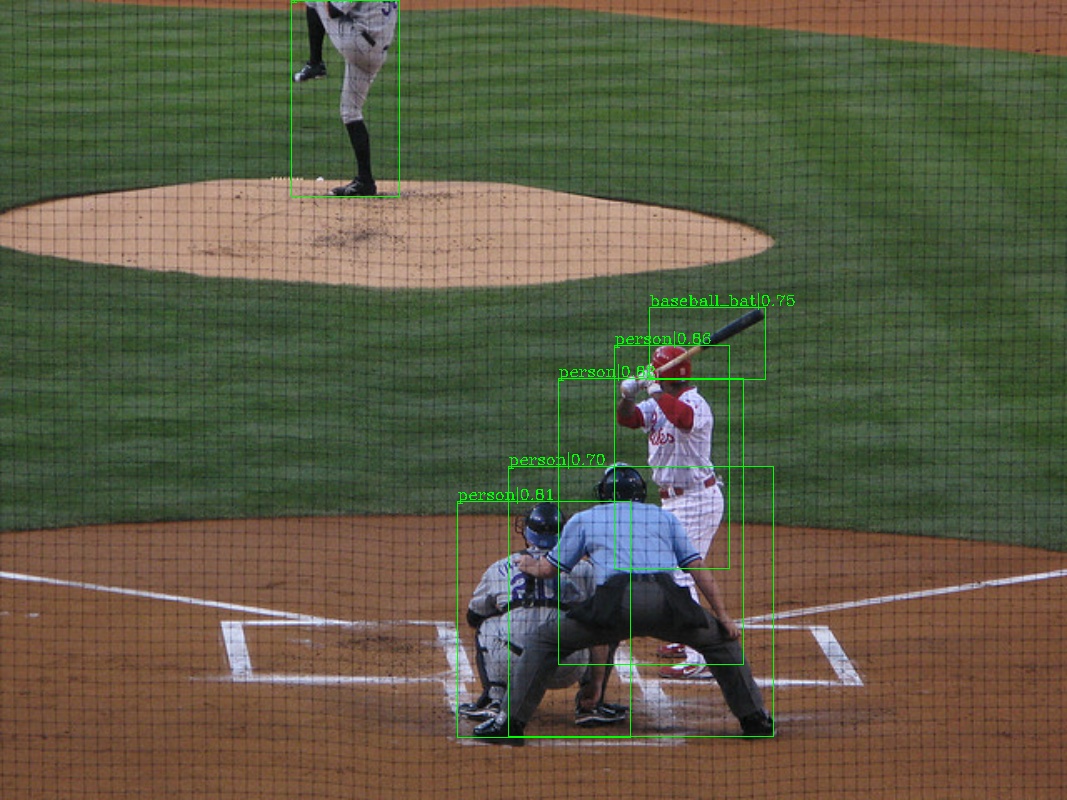}}
\subfigure{
\centering
\includegraphics[width=0.191\textwidth,height=0.11\textheight]{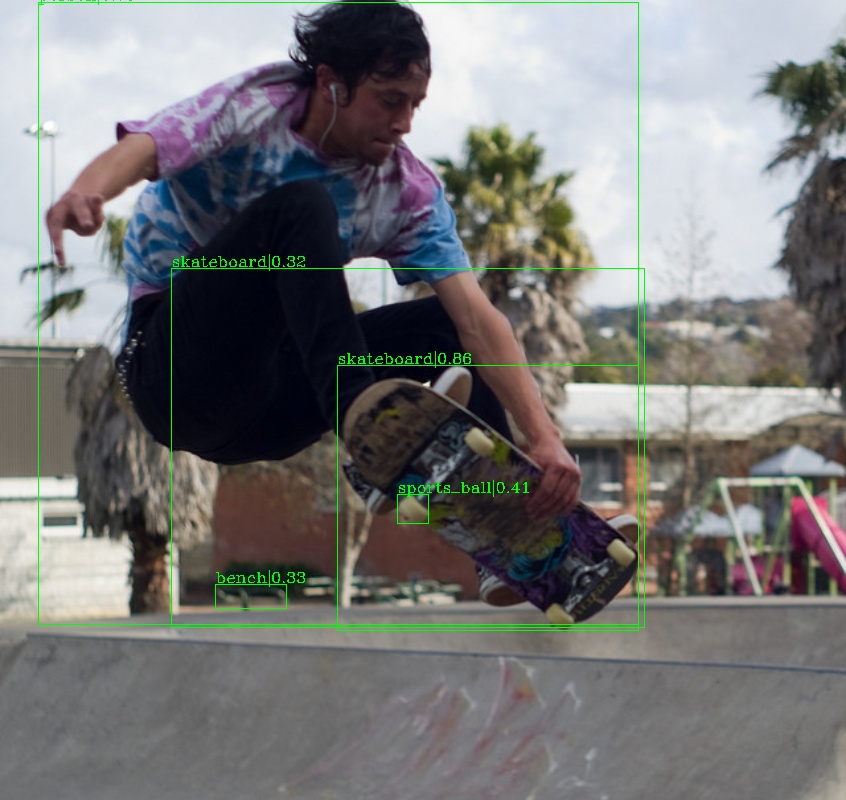}}
\vspace{-.10in}
\subfigure{
\centering
\includegraphics[width=0.191\textwidth,height=0.11\textheight]{4_baseline.jpg}}
\subfigure{
\centering
\includegraphics[width=0.191\textwidth,height=0.11\textheight]{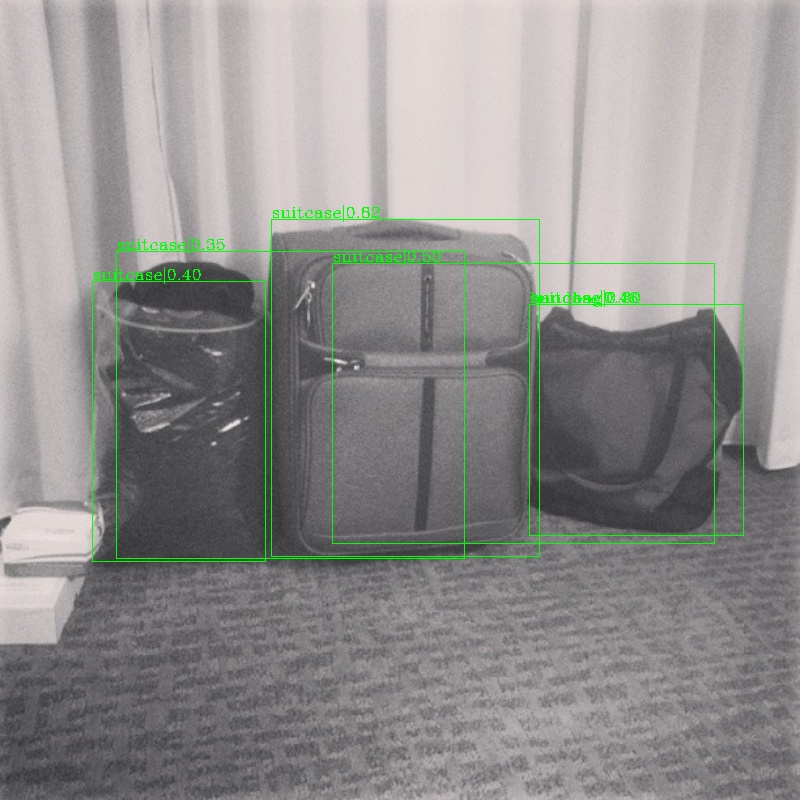}}
\vspace{-.10in}
\subfigure{
\centering
\includegraphics[width=0.191\textwidth,height=0.11\textheight]{10.jpg}}
\subfigure{
\centering
\includegraphics[width=0.191\textwidth,height=0.11\textheight]{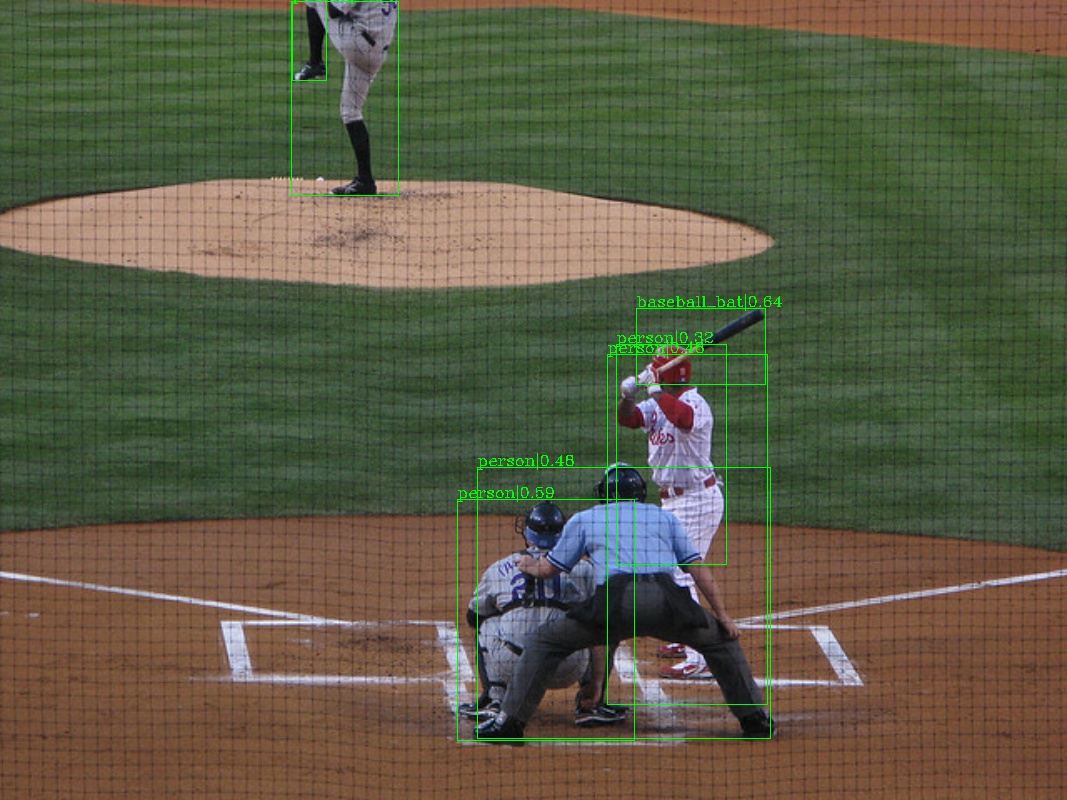}}
\subfigure{
\centering
\includegraphics[width=0.191\textwidth,height=0.11\textheight]{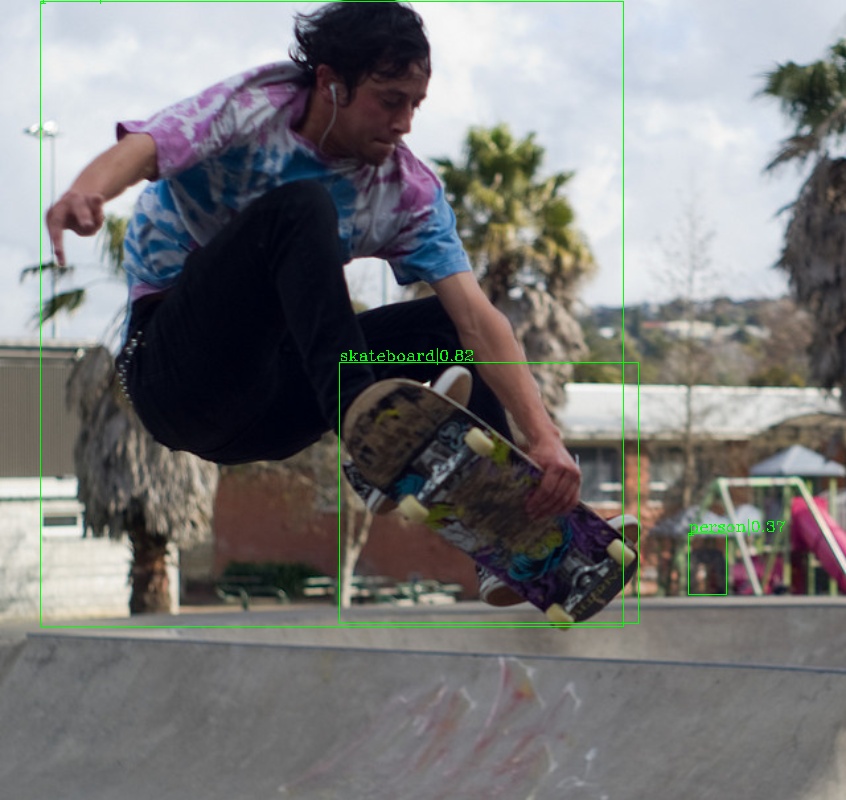}}
\subfigure{
\centering
\includegraphics[width=0.191\textwidth,height=0.11\textheight]{4.jpg}}
\subfigure{
\centering
\includegraphics[width=0.191\textwidth,height=0.11\textheight]{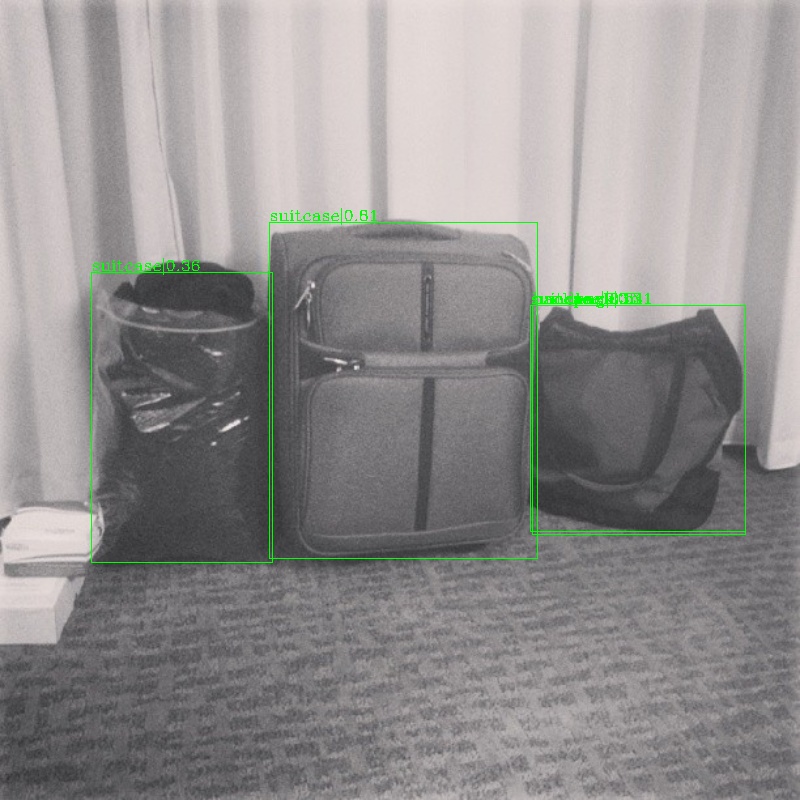}}
\vspace{-.10in}
\subfigure{
\centering
\includegraphics[width=0.191\textwidth,height=0.21\textheight]{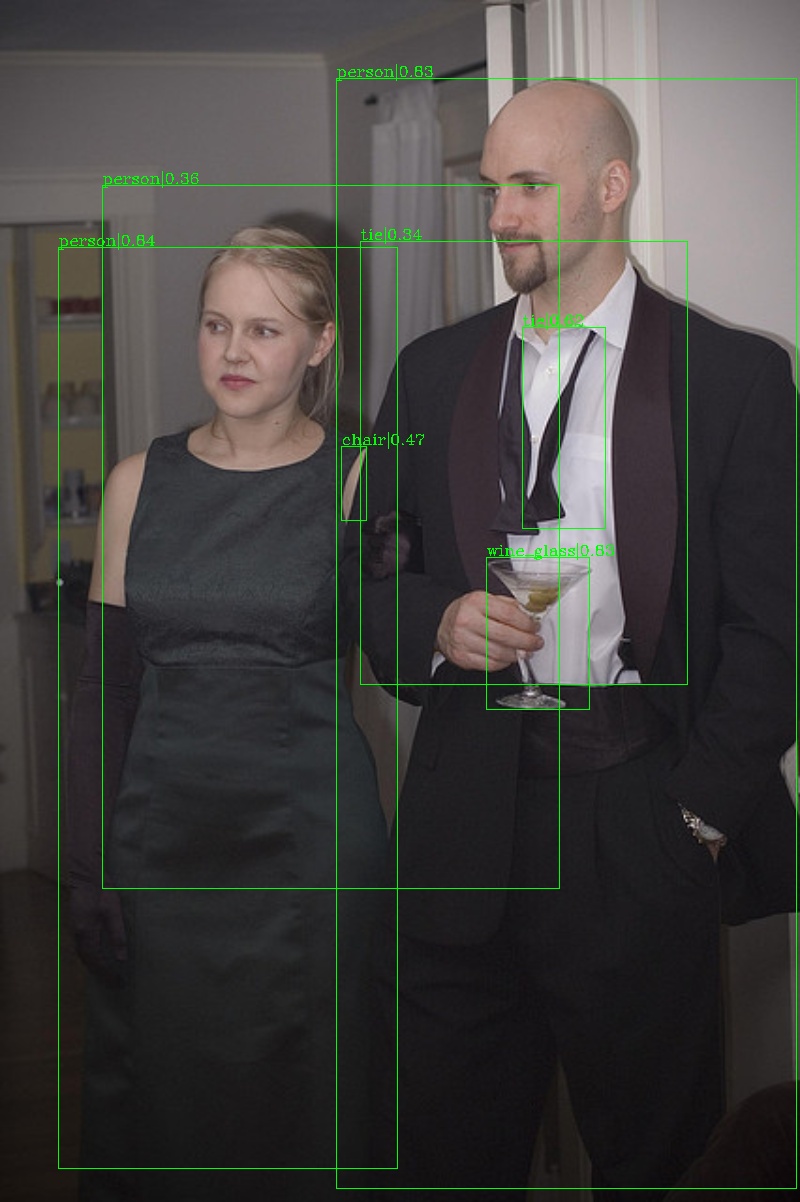}}
\subfigure{
\centering
\includegraphics[width=0.191\textwidth,height=0.21\textheight]{8_baseline.jpg}}
\subfigure{
\centering
\includegraphics[width=0.191\textwidth,height=0.21\textheight]{13_baseline.jpg}}
\subfigure{
\centering
\includegraphics[width=0.191\textwidth,height=0.21\textheight]{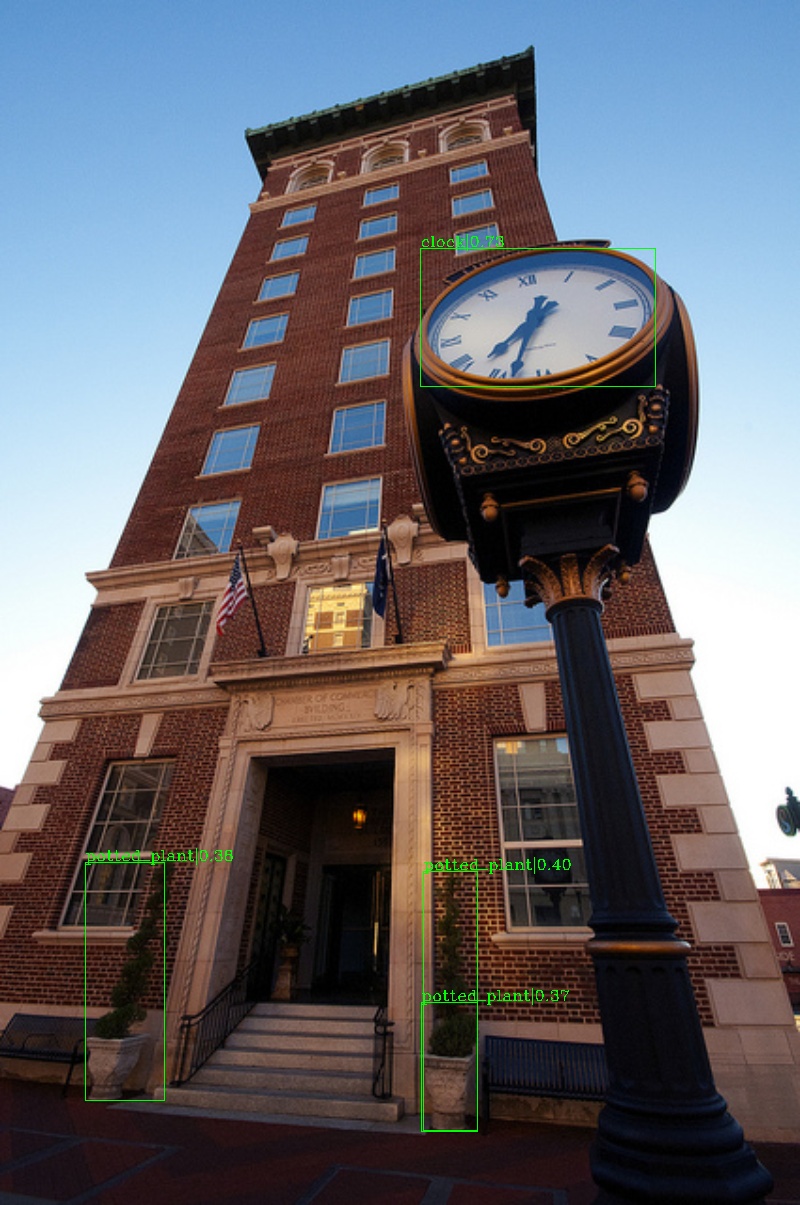}}
\subfigure{
\centering
\includegraphics[width=0.191\textwidth,height=0.21\textheight]{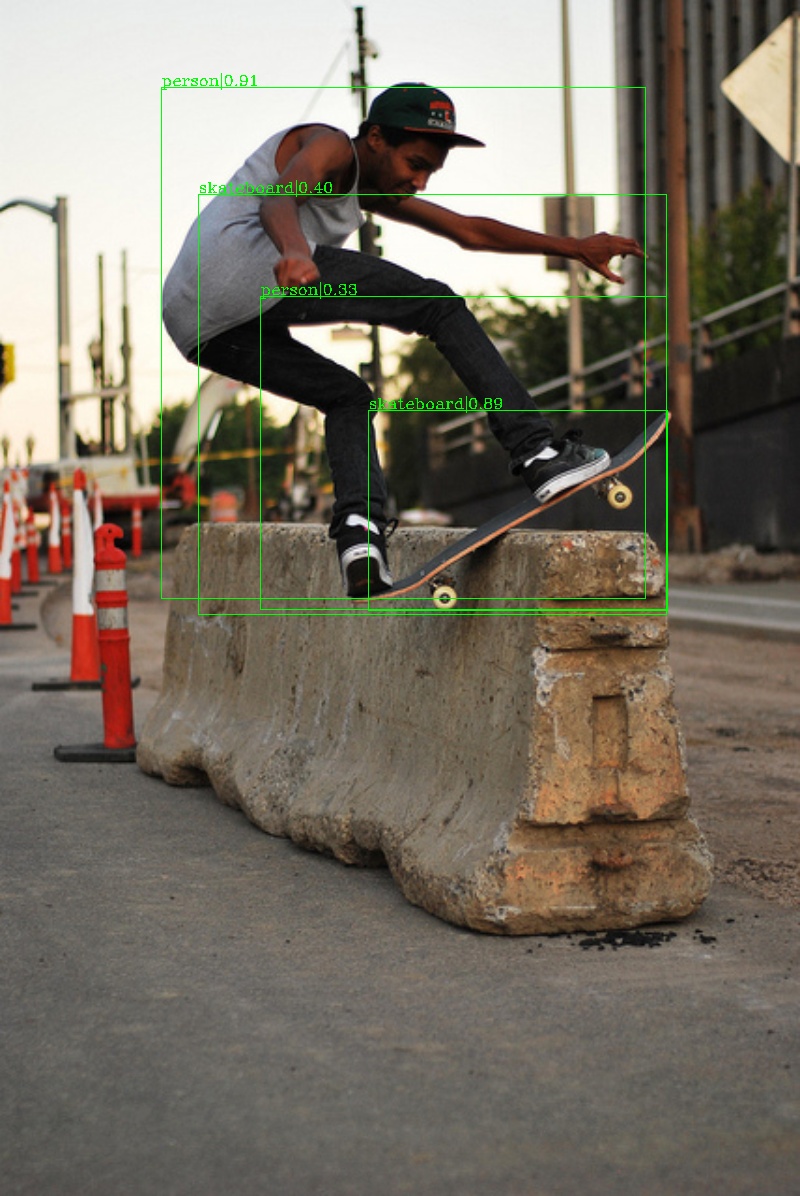}}
\vspace{-.10in}
\subfigure{
\centering
\includegraphics[width=0.191\textwidth,height=0.21\textheight]{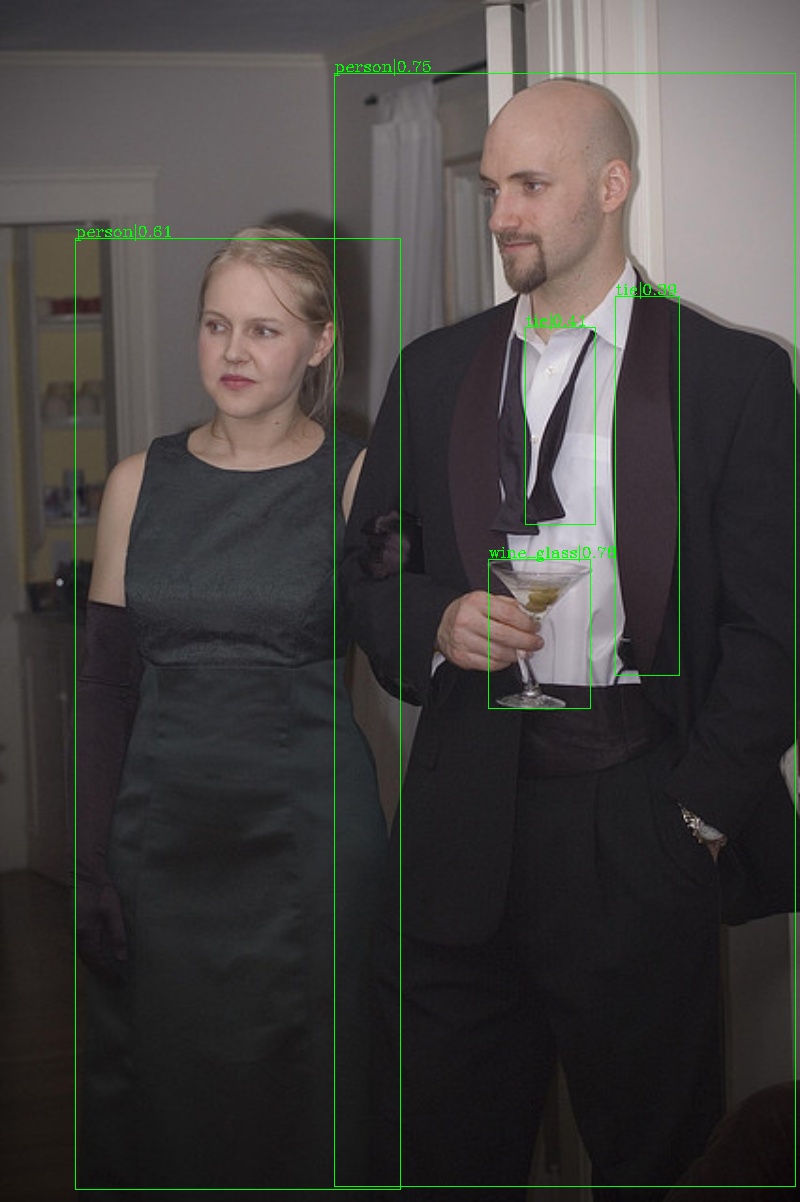}}
\subfigure{
\centering
\includegraphics[width=0.191\textwidth,height=0.21\textheight]{8.jpg}}
\subfigure{
\centering
\includegraphics[width=0.191\textwidth,height=0.21\textheight]{13.jpg}}
\subfigure{
\centering
\includegraphics[width=0.191\textwidth,height=0.21\textheight]{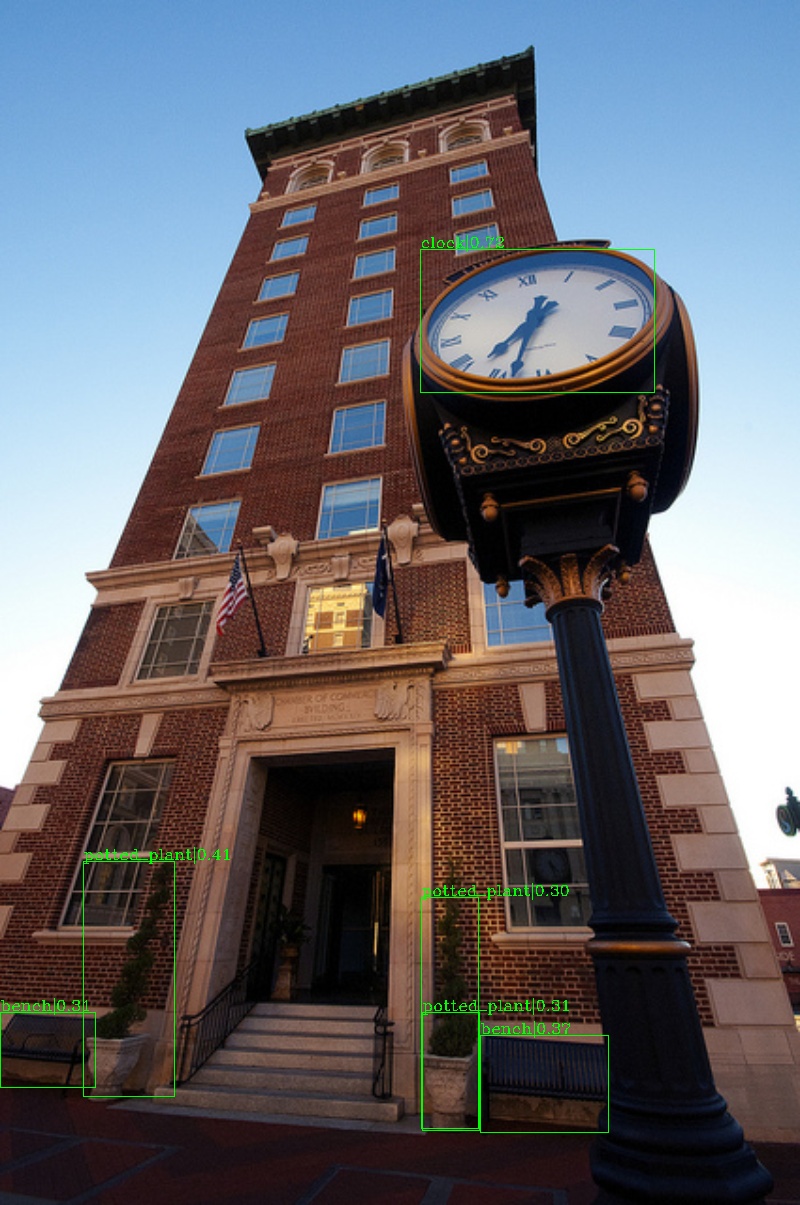}}
\subfigure{
\centering
\includegraphics[width=0.191\textwidth,height=0.21\textheight]{14.jpg}}

\caption{Visualization of some detection results that are best viewed when zoomed in. The first and third rows of images are detection results of anchor-free baseline; the second and fourth rows of images are detection results of anchor-free baseline with sampling based on PIoU metric at 0.5 threshold. As shown from the detection results, our method produces much less false positives and more accurate localization.}
\label{Figure 4}
\end{figure*}

\subsection{Comparison with State-of-the-art Detectors }
\noindent \textbf{MSCOCO minival set}~
Firstly, we validate our model on MSCOCO minival set. The baseline model employs ResNet-50-FPN backbone with fixed batch normalization layers and an anchor-free detection head that takes all points from feature pyramids inside ground truth as positive samples. The input image is resized to their shorter side being 800 and their longer side less or equal to 1333 for single-scale training. Image flipping is the only strategy for data augmentation. During testing, we only use the same scale setting to the training process and all results are listed in Table~\ref{Table 3}. Without accurate assignment, there is a performance gap between our anchor-free baseline and RetinaNet. After using Pseudo-IoU based assigment, it outperforms RetinaNet about $1.1\%$ mAP performance and brings $2.9\%$ mAP improvement on baseline model. Moreover, employing GIoU and centerness branch could further improve the performance, which implies that our Pseudo-IoU is compatible with other object detection approaches.
\\
\\
\noindent \textbf{MSCOCO test-dev set}~ We also report our results on test-dev set. We train all model with a ResNet-101-FPN backbone and an anchor-free detection head illutrated in Figure~\ref{Figure 3}. The input image is resized to keep the longer side to 1333 and short side randomly selected from 640 to 800 without image flipping or other data augmentation approach. All detection results are listed in Table~\ref{Table 4}. To make fair comparisons, we only select results with single-model and single-scale inference from their original publications, some results based on multi-scale testing or better backbone like ResNeXt~\cite{xie2017aggregated} are not included.

\subsection{Visualization}
We visualize some detection results of both anchor-free baseline and the baseline trained with sampling in Figure~\ref{Figure 4}. It is noticeable that the baseline model has many false positives and inaccurate bounding box predictions. On the contrary, the model adopts sampling based on PIoU produces much less false positives and more accurate localization, which improves performance by a large margin.

\section{Conclusion}
In this paper, we first point out that current anchor-free object detectors lack accurate and standardized assignment process that limits its potential to compete with many state-of-the-art anchor-based methods. To solve this problem, we propose a simple Pseudo-Intersection-over-Union~(Pseudo-IoU) metric that brings more accurate and standardized assignment rule into anchor-free framework by computing IoU between the pseudo box centered on points from feature pyramids and ground truth box. Then, we adopt it as Pseudo-IoU value of each point inside ground truth area. Furthermore, we conduct extensive experiments on PASCAL VOC and MSCOCO dataset, it shows that anchor-free models with Pseudo-IoU based assignment could bring a consistent improvement without bells and whistles. It decreases many false positive detections and leads to more accurate localization of bounding boxes. Moreover, in the future, it is possible that some recent research on improving assignment in anchor-based methods can be transferred to anchor-free models based on the proposed Pseudo-IoU assignment metric.

\section*{Acknowledgement}
This work is supported by IBM-ILLINOIS Center for Cognitive Computing Systems Research (C3SR) - a research collaboration
as part of the IBM AI Horizons Network. We gratefully acknowledge the
support and advice from C3SR, with providing computational GPU resources for the experiments of this research.

{\small
\bibliographystyle{ieee_fullname}
\bibliography{egbib}
}

\end{document}